\documentclass[10pt,twocolumn,letterpaper]{article}

\usepackage[pagenumbers]{cvpr} 

\newtheorem{theorem}{Theorem}[section]

\newtheorem{lemma}[theorem]{Lemma}

\usepackage{wrapfig}
\usepackage{tabularx}
\usepackage{xcolor}
\usepackage{hyperref}
\usepackage{makecell}

\title{Goal-Guided Reinforcement Learning via Fine-Tuning Vision-Language Models}

\author{
Yajie Qi$^1$,\quad
Wei Wei$^1$,\quad
Lin Li$^1$,\quad
Lijun Zhang$^1$,\quad
Jianye Hao$^2$
\\
$^1$ Key Laboratory of Computational Intelligence and Chinese Information Processing of Ministry \\
of Education, School of Computer and Information Technology, Shanxi University \\  
$^2$ College of Intelligence and Computing, Tianjin University
}

\begin{document}
\maketitle

\begin{abstract}
Exploration remains a major challenge for reinforcement learning (RL) agents in Craftax-style environments due to sparse rewards and long horizons.
Large language models (LLMs) can guide exploration by generating high-level goals, but methods using models without explicit task adaptation often rely on carefully designed prompts and may produce inconsistent or contextually inappropriate goals.
We introduce CCGP-2M, a multimodal goal-preference dataset constructed from interaction trajectories in Craftax-Classic, and propose a vision-language model (VLM)-guided RL framework that learns environment-specific goal guidance from this dataset. 
The framework consists of two stages: 
offline VLM fine-tuning and online goal-guided RL. In the offline stage, we align the VLM with goal preferences using Kullback-Leibler Token-level Direct Preference Optimization (KLTDPO), which combines token-level KL regularization with distributionally robust reweighting. 
Experiments on Craftax-Classic show that our method collects diamonds earlier than state-of-the-art baselines, achieves the highest score and reward, and attains the highest success rates on most achievements.
\end{abstract}

\section{Introduction}
\label{sec:intro}

Reinforcement learning (RL) has achieved remarkable success in a wide range of sequential decision-making tasks~\citep{mnih2015human, shakya2023reinforcement, oh2025discovering}.
However, these successes have largely been achieved in closed environments characterized by predefined tasks, explicit goals, and immediate feedback, whereas efficient exploration remains challenging in open-world environments such as Minecraft~\citep{fan2022minedojo,li2025open} and Craftax-Classic~\citep{matthews2024craftax,simmons2025deep}.
In these environments, sparse rewards and long horizons make trial-and-error exploration inefficient.
Recently, LLM-guided RL has emerged as a promising method for addressing this challenge, driven by the remarkable capabilities of LLMs in complex decision-making and reasoning tasks~\citep{Zhou2024LLM4RL,schoepp2025evolving}.

\begin{figure}[t]
   \centering
   \includegraphics[width=0.98\linewidth]{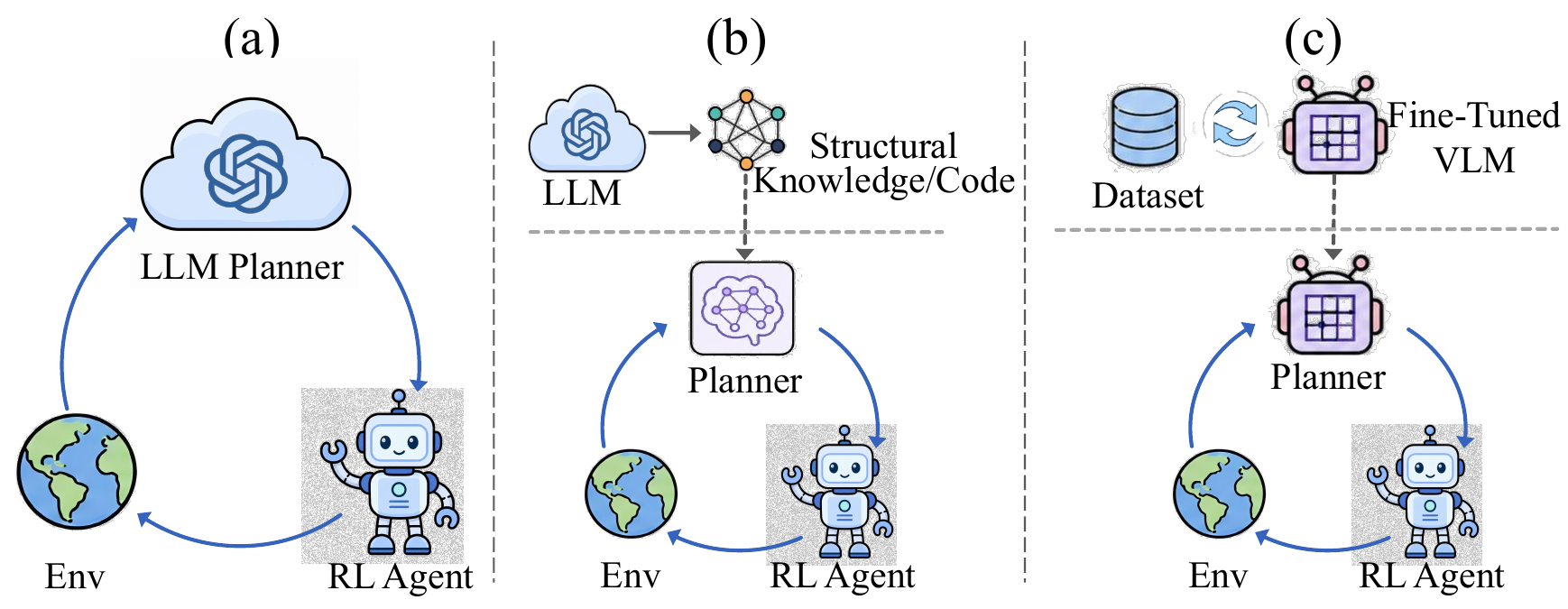}
   \caption{Comparison of existing methods and our framework.
   Prior work either 
   (a) directly uses cloud-based LLM planners for RL guidance or 
   (b) enhances local planners with LLM-derived structured knowledge or code.
   (c) Our framework fine-tunes a local VLM planner with goal-preference data to provide RL guidance.}
   \label{fig:motivation}
\end{figure}

Existing LLM-guided RL methods mainly follow two categories. 
The first category, illustrated in Fig.~\ref{fig:motivation}(a), directly queries LLMs to generate goals that guide RL exploration~\citep{du2023guiding,zhang2024adarefiner,fu2026counterfactual}. 
These methods often rely on models that have not been explicitly adapted to the target environment. 
To obtain useful guidance, they must convey environment rules, agent states, and task requirements through carefully designed prompts. 
Even with such prompts, the generated goals may be inconsistent across queries or poorly matched to the current state. 
Frequent queries to large-scale or commercial models can also introduce latency and financial costs. 
The second category uses LLMs to extract reusable structured task knowledge or generate executable planning code for local planners~\citep{chen2025causal,fan2025subgoal,qi2025goal}, as illustrated in Fig.~\ref{fig:motivation}(b).
By moving knowledge extraction or code generation offline, these methods reduce reliance on frequent LLM queries. 
However, constructing useful knowledge or planning code still requires task-specific prompts, and the resulting guidance depends on the quality of the LLM outputs. 
Errors or omissions in these outputs may lead to impractical goals during exploration.

These limitations motivate an alternative that learns environment-specific goal guidance from interaction data.
As illustrated in Fig.~\ref{fig:motivation}(c), we construct CCGP-2M, a multimodal goal preference dataset derived from environment interaction trajectories, and use it to fine-tune a local VLM that generates goals to guide RL exploration.
Our main contributions are summarized as follows:
\begin{itemize}
    \item We construct a large-scale multimodal dataset, called the Craftax-Classic Goal Preference dataset (CCGP-2M), derived from goal-conditioned interaction trajectories without manual annotation. CCGP-2M contains 2 million preference samples with 43 distinct goals, covering all 22 achievements, diverse environment states and task stages in Craftax-Classic.
   
    \item We propose a two-stage VLM-guided RL framework consisting of offline VLM fine-tuning and online goal-guided RL. In the offline stage, we introduce KLTDPO, which integrates token-level sequential KL regularization with distributionally robust reweighting to fine-tune a lightweight VLM on CCGP-2M dataset. In the online stage, the fine-tuned VLM serves as a planner and generates high-level goals to guide RL exploration.

    \item We evaluate our method on Craftax-Classic across all 22 achievements. Compared with state-of-the-art baselines, our method collects diamonds earlier, achieves higher score and reward, and ranks first in success rate on 15 achievements and second on the remaining achievements.
\end{itemize}

\section{Preliminaries}
\label{sec:formatting}

\subsection{Goal-Conditioned POMDP}

A goal-conditioned partially observable Markov decision process (GC-POMDP) is defined by  $\mathcal{M}=(\mathcal{S},\mathcal{G},\mathcal{A},\Omega,P,\mathcal{O},R,\gamma)$~\citep{zhao2023decision}, where $\mathcal{S}$, $\mathcal{G}$, $\mathcal{A}$, and $\Omega$ denote the state, goal, action, and observation spaces, respectively.
The transition kernel $P:\mathcal{S}\times\mathcal{A}\rightarrow\Delta(\mathcal{S})$ governs the environment dynamics, while the observation kernel $\mathcal{O}:\mathcal{S}\times\mathcal{A}\rightarrow\Delta(\Omega)$ generates the agent's partial observations.

At each time step $t$, the agent's observation is $o_t\in\Omega$, and the goal is $g\in\mathcal{G}$. The agent takes an action $a_t\in\mathcal{A}$ according to the policy $\pi_\theta(\cdot\mid o_t,g)$, causing the environment to transition to a new state $s_{t+1}$ according to $P(s_{t+1}\mid s_t,a_t)$. The agent aims to maximize the expected sum of discounted rewards under goal $g$:
\begin{equation}
    J(\pi_\theta)=\mathbb{E}_{\pi_\theta}\left[\sum_{t=0}^{\infty}\gamma^t r_t(s_t,a_t)\right].
    \label{eq:pomdp_objective}
\end{equation}

\begin{figure*}
    \centering
    \includegraphics[width=0.98\linewidth]{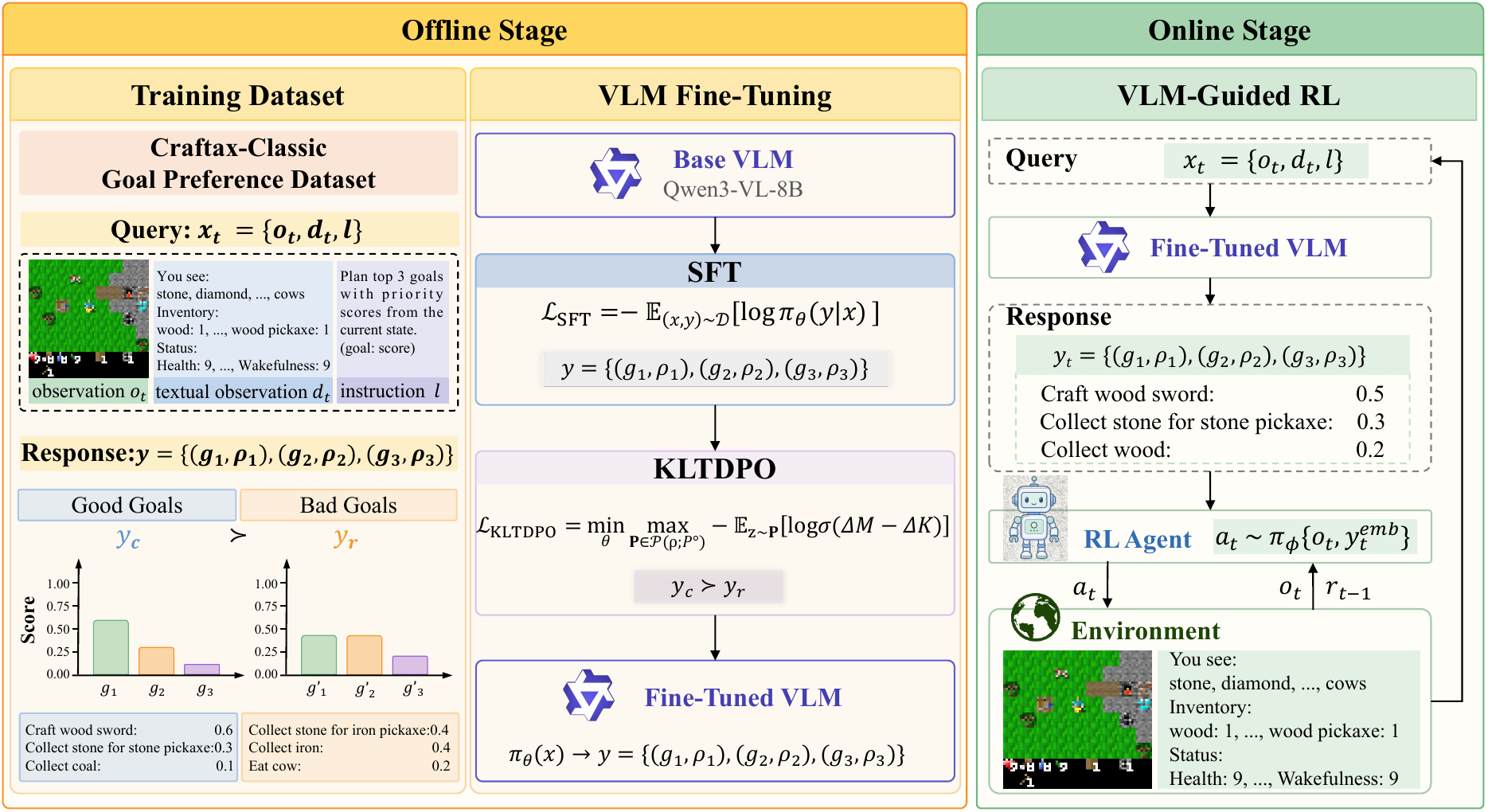} 
    \caption{
    Overview of our framework.
    }
    \label{fig:framework}
\end{figure*}

\subsection{Direct Preference Optimization}

Direct Preference Optimization (DPO)~\citep{rafailov2023direct_dpo} aligns a policy by comparing pairwise preferences without explicitly learning a reward model.
Specifically, starting from the following KL-regularized RL objective:
\begin{equation} 
    \begin{aligned}
        \max_{\pi_{\theta}} \mathbb{E}_{x \sim \mathcal{D}} & [\mathbb{E}_{y \sim \pi_{\theta}(\cdot \mid x)}[r \left(x, y\right) \\
        & -\beta D_{\mathrm{KL}}\left(\pi_{\theta}(\cdot \mid x) \| \pi_{\mathrm{ref}}(\cdot \mid x)\right)]],
    \end{aligned}
    \label{Equation: RL objective}
\end{equation}
where $\mathcal{D}$ is the preference dataset, $r^{\ast}(x,y)$ is the underlying reward for response $y$ to prompt $x$, $\pi_{\theta}$ is the policy to be optimized, and $\pi_{\mathrm{ref}}$ is a fixed reference policy, typically chosen as a supervised fine-tuned model.
The coefficient $\beta > 0$ controls the strength of the reverse KL penalty.

Preference probabilities are modeled using the Bradley-Terry (BT) pairwise comparison~\citep{1952_bradley_BT-model}:
\begin{equation}
    P_{\mathrm{BT}}\left(y_c \succ y_r \mid x\right) = \frac{\exp\left(r^{\ast}\left(x, y_c\right)\right)}{\exp\left(r^{\ast}\left(x, y_c\right)\right) + \exp \left(r^{\ast}\left(x, y_r\right)\right)},
    \label{eq:BT}
\end{equation}
where $y_c \succ y_r \mid x$ indicates that the chosen response $y_c$ is preferred over the rejected response $y_r$ for prompt $x$.

DPO  establishes a direct functional mapping between the reward model and the optimal policy:
\begin{equation}
     r^{\ast}(x, y)=\beta \log \frac{\pi^{\ast}_{\theta}(y \mid x)}{\pi_{\mathrm{ref}}(y \mid x)}+\beta \log Z(x),
    \label{eq:DPO reward}
\end{equation}
where $\pi^{\ast}_{\theta}$ denotes the optimal policy and $Z(x)$ is the corresponding partition function.

Substituting this mapping into the BT model yields the DPO preference loss for a parameterized policy $\pi_{\theta}$:
\begin{equation}
    \mathcal{L}_{\mathrm{DPO}}\left(\pi_{\theta}; \pi_{\mathrm{ref}}\right)=-\mathbb{E}_{\left(x, y_c, y_r\right) \sim \mathcal{D}}\left[\log\sigma\left(\Delta M\right)\right],
\end{equation}
where $\sigma$ is the sigmoid function and $\Delta M$ is the difference between the implicit rewards assigned by $\pi_{\theta}$ relative to $\pi_{\mathrm{ref}}$:
\begin{equation}
\Delta  M = \beta\log\frac{\pi_{\theta}\left(y_c \mid x\right)}{\pi_{\mathrm{ref}}\left(y_c \mid x\right)} - \beta\log\frac{\pi_{\theta}\left(y_r \mid x\right)}{\pi_{\mathrm{ref}}\left(y_r \mid x\right)}.
\label{eq:delta_m}
\end{equation}

Minimizing this objective increases the relative likelihood of chosen responses while maintaining proximity to the reference policy.
Building upon DPO, TDPO~\citep{zeng2024token} introduces token-level KL regularization, while KLDPO~\citep{xu2025robust} improves preference optimization through robust loss weighting.
Details of these variants are provided in Appendix~A.

\section{Method}

In this section, we develop a novel VLM-guided RL approach. 
Specifically, we first construct CCGP-2M, a multimodal goal preference dataset containing 2 million multimodal preference samples and covering all 22 achievements, a broad range of environment states, and diverse task goals. 
Second, as illustrated in Fig.~\ref{fig:framework}, we propose a two-stage VLM-guided RL framework that combines offline VLM fine-tuning and online goal-guided RL.

\subsection{Dataset Construction} \label{sec:dataset_construction}

In this subsection, we construct CCGP-2M, a multimodal goal preference dataset derived from goal-conditioned interaction trajectories without manual goal annotations. 
Its construction faces three challenges: 
(1) difficult achievements require long prerequisite behaviors and are hard to collect; 
(2) raw trajectories contain redundant observations and repeated goals; 
and (3) preference learning requires structurally valid yet contextually inappropriate rejected responses. 
To address these challenges, we collect diverse trajectories under random and partial initialization, apply quality-aware sample selection to retain informative samples, and construct preference pairs by contrasting trajectory-consistent plans with carefully designed rejected alternatives.

\paragraph{Trajectory Collection.}

\begin{table}[t]
\centering
\small
\setlength{\tabcolsep}{4pt}
\caption{
    Trajectory collection statistics under random and partial initialization.
}
\begin{tabular}{lrrr}
\toprule
\textbf{Statistic} &
\textbf{Random} &
\textbf{partial} &
\textbf{Total} \\
\midrule
Environment steps 
& $60$M & $30$M & $90$M \\
Trajectories 
& $27{,}355$ & $360$ & $27{,}715$ \\
Observations 
& $8{,}858{,}665$ & $77{,}816$ & $8{,}936{,}481$ \\
\midrule
Observations with diamond 
& $12{,}675$ & $41{,}608$ & $54{,}283$ \\
\bottomrule
\end{tabular}
\label{tab:trajectory_statistics}
\end{table}

Following the implementation of SGRL~\citep{qi2025goal}, we collect goal-conditioned interaction trajectories in the Craftax-Classic environment, where the exploration goals for the RL agent are specified by planning codes generated by DeepSeek-R1~\citep{guo2025deepseek} and Claude-Opus-4.7~\citep{anthropic2026claudeopus47}.
Each collected trajectory is represented as $\tau_i=\{(o_t,d_t,y_t,r_t,score_t)\}_{t=1}^{T_i}$, where $o_t$ is the observation, $d_t$ is its textual description, $y_t$ is the goal set, $r_t$ is the reward, and $score_t$ is the score at time step $t$. Here, $T_i$ denotes the length of the $i$-th trajectory.
Notably, we employ two environment initialization strategies to increase trajectory diversity and ensure sufficient coverage of states that enable the completion of deep achievements. 
Specifically, we use random initialization together with a manually specified setting in which the agent starts with an iron pickaxe to facilitate diamond-related exploration. 
Under these two settings, we collect 60M and 30M environment steps, respectively, and retain approximately 27,000 and 360 valid trajectories, yielding a dataset containing over 8.9 million observations. 
Statistics of the collected goal-conditioned interaction trajectories are summarized in Tab.~\ref{tab:trajectory_statistics}.
More details are provided in Appendix~B.1.

\paragraph{Quality-Aware Sample Selection.}

Raw trajectories contain redundant and low-quality records, such as repeated observations, malformed planner outputs, and persistent local goals, which may bias planner fine-tuning toward frequent but less informative behaviors.

Invalid and duplicated records are first removed according to observation validity, goal format consistency, priority constraints, and normalized observation-goal uniqueness.
For each remaining candidate at step $t$, we compute a quality score
$
S^{\text{quality}}_t=B_t\min(R_t,R_{\max})P^{\text{goal}}_t+\epsilon_t,
$
where $B_t$ captures informative trajectory events, including goal transitions, achievement proximity, state changes, and survival needs; $R_t$ is a frequency-based rarity multiplier; $R_{\max}$ controls the maximum contribution of goal rarity; $P^{\text{goal}}_t$ penalizes recently repeated goals; and $\epsilon_t$ breaks ties between candidates with identical deterministic scores.

To avoid overrepresenting long trajectories while preserving the empirical state distribution, each episode is assigned a sampling quota
\begin{equation}
M_i = \left\lfloor \mathcal{T} \left(\lambda\frac{1}{N_{\mathrm{ep}}} + (1-\lambda)\frac{N_i}{\sum_jN_j} \right) \right\rfloor ,
\label{eq:dataset_episode_quota}
\end{equation}
where $M_i$ denotes the sampling quota assigned to episode $i$, $\mathcal{T} $ represents the total sampling budget, and $N_{\mathrm{ep}}$ is the total number of episodes.
$N_i$ denotes the number of valid candidates extracted from episode $i$.
The coefficient $\lambda\in[0,1]$ controls the trade-off between uniform episode-level coverage and candidate-proportional allocation.
Then, all valid candidates within episode $i$ are ranked according to $S^{\text{quality}}_t$ and the top $\min(M_i,N_i)$ candidates are retained.
We set $R_{\max}=4.0$ and sample $\epsilon_t$ from $\mathcal{U}(0,0.15)$ only for tie breaking.
The exact signal weights and decision rules are reported in Appendix~B.2.

\paragraph{Preference Pair Construction.}

\begin{figure}[t]
   \centering
   \includegraphics[width=0.9\linewidth]{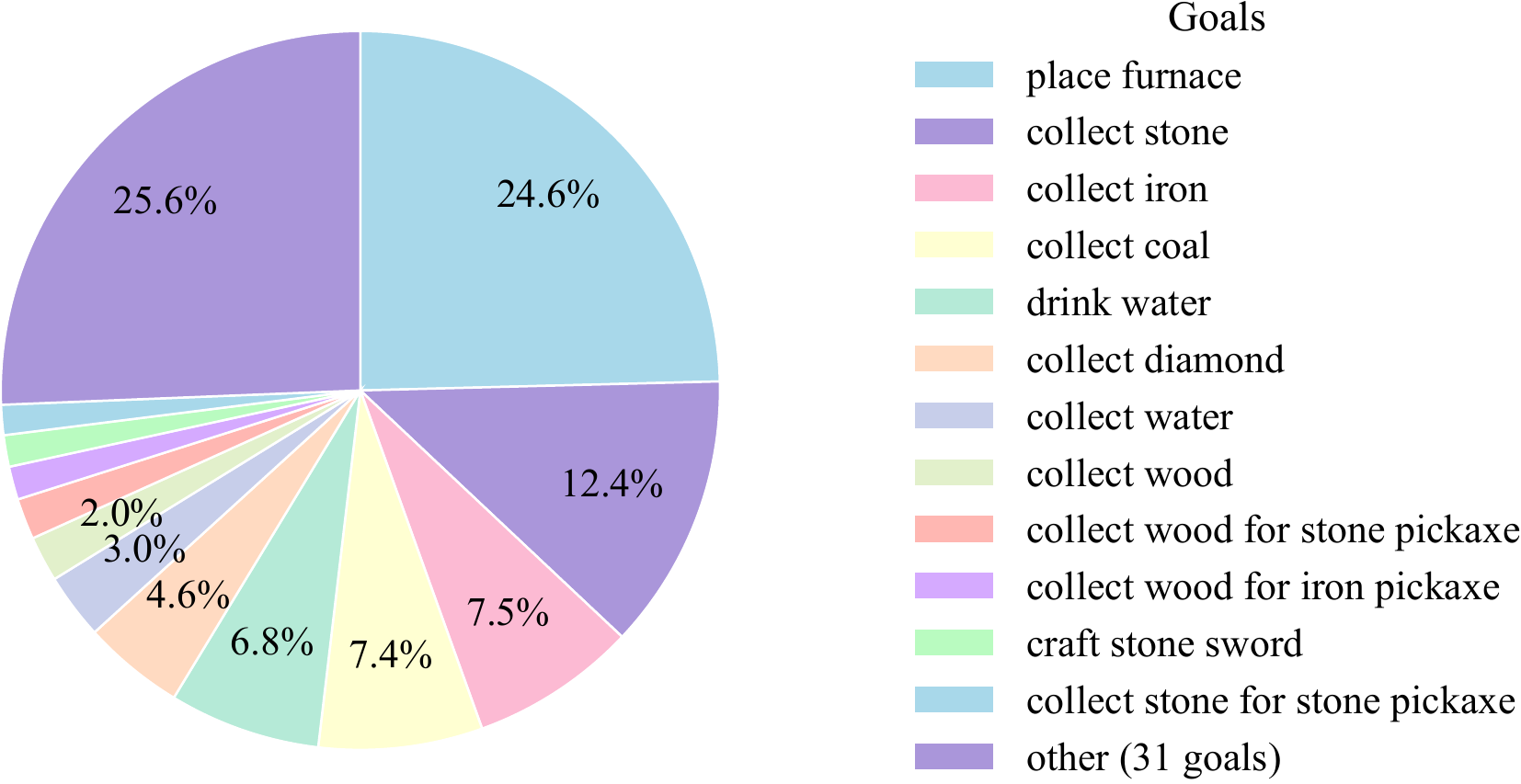}
   \caption{Distribution of the goals in CCGP-2M preference dataset.}
   \label{fig:goal_distribution}
\end{figure}

For each retained query $x$ at trajectory step $t$, we use the recorded planner response $y_t$ as the chosen response $y_c$.
A rejected response $y_r$ is then constructed to preserve the required output format while being less suitable for the current state or task stage.
Specifically, $y_r$ is sampled from a mixture of context-dependent negative construction strategies:
\begin{equation}
y_r \sim q_{\mathrm{neg}}(\cdot\mid\tau,t,y_c),\label{eq:dataset_negative_distribution}
\end{equation}
where
$
q_{\mathrm{neg}}= \lambda_f q_{\mathrm{future}}+\lambda_p q_{\mathrm{past}}+\lambda_m q_{\mathrm{mixed}} +\lambda_s q_{\mathrm{priority}}+\lambda_g q_{\mathrm{global}},
$
the nonnegative weights satisfy
$\lambda_f+\lambda_p+\lambda_m+\lambda_s+\lambda_g=1$.
Here, $q_{\mathrm{future}}$ and $q_{\mathrm{past}}$ sample responses from nearby future and past steps to create premature or obsolete plans.
$q_{\mathrm{mixed}}$ replaces part of $y_c$ with subgoals from neighboring steps,
$q_{\mathrm{priority}}$ changes goal ordering or priority scores while preserving the goal set, and $q_{\mathrm{global}}$ samples a different response from the same trajectory.
These strategies produce hard negatives that remain fluent and structurally valid but differ from $y_c$ in temporal relevance, goal composition, or priority assignment.
Then, the resulting preference dataset is named CCGP-2M, which is denoted by 
\begin{equation}
    \mathcal{D}_{\mathrm{pref}} = \left\{(x^i,y_c^i,y_r^i) \right\}_{i=1}^{N},
\end{equation}
where $y_c^i \succ y_r^i$.
The CCGP-2M dataset contains $2$ million preference samples covering $43$ distinct goals.
The distribution of goal types in CCGP-2M is illustrated in Fig.~\ref{fig:goal_distribution}.

In practice, for SFT, we ignore the rejected responses and use only the preferred pairs:
$(x,y_c)\sim\mathcal{D}_{\mathrm{pref}}^{+},$
where $\mathcal{D}_{\mathrm{pref}}^{+}$ denotes the positive subset extracted from $\mathcal{D}_{\mathrm{pref}}$.
For preference optimization, the complete preference samples in $\mathcal{D}_{\mathrm{pref}}$ are used to optimize the preference objective.
More details are provided in Appendix~B.3.

\begin{figure*}
\centering
\includegraphics[width=0.9\linewidth]{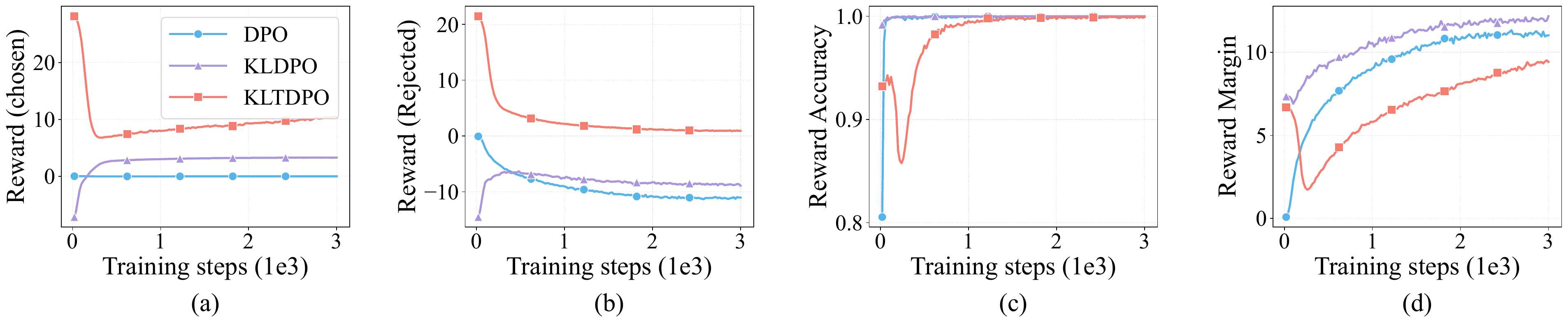} 
\caption{The reward curves for VLMs fine-tuned with DPO, KLDPO, and KLTDPO.}
\label{fig:preference_optimization_analysis}
\end{figure*}

\subsection{VLM-guided RL framework}

The VLM-guided RL framework consists of two stages: offline VLM fine-tuning and online goal-guided RL.
In the offline stage, we introduce KLTDPO, which integrates token-level sequential KL regularization with distributionally robust reweighting to fine-tune a lightweight VLM on CCGP-2M dataset. 
In the online stage, the fine-tuned VLM serves as a planner and generates high-level goals to guide RL exploration.

\paragraph{VLM Fine-tuning.}

In this subsection, we focus on how to further fine-tune the VLM model that has undergone SFT with the loss
\begin{equation}
    \mathcal{L}_{\mathrm{SFT}} = -\mathbb{E}_{(x,y_c)\sim\mathcal{D}_{\mathrm{pref}}^{+}}\left[\sum_{t=1}^{|y_c|}\log\pi_{\theta}\left(y^t_c\mid x,y_c^{<t}\right)\right],
\end{equation}
using the CCGP-2M preference dataset.

Specifically, we first use DPO and KLDPO to fine-tune the VLM and find thatboth methods rapidly distinguish chosen responses from rejected ones, as shown by their increasing reward margins and reward accuracies that approach 1 in Fig.~\ref{fig:preference_optimization_analysis}. 
However, when the fine-tuned VLMs are used as planners to guide RL exploration, the performance gains of the RL agent remain limited, as shown in Tab.~\ref{tab:main_results}.
We hypothesize that this may be attributed to the distinctive data distribution of CCGP-2M. 
Specifically, the chosen response is a trajectory-consistent plan, whereas the rejected responses are diverse but often constitute easy negatives generated by modifying the goals, their ordering, or their priority scores.

Motivated by TDPO~\citep{zeng2024token}, we propose KLTDPO, which combines token-level sequential KL regularization with the distributionally robust reweighting mechanism of KLDPO~\citep{xu2025robust} to emphasize difficult preference samples and enable finer-grained preference optimization for planning decisions.
The detailed training procedure is as follows.

\paragraph{Training procedure.}

We first formulate autoregressive response generation as an augmented preference-based MDP, where the state at time step $t$ includes the prompt $x$ and the first $t-1$ tokens $y^{<t}$ of the response, denoted as $s_t = [x, y^{<t}]$,
the action corresponds to the next generated token, denoted as $a_t = y^t$,
and the reward function $r(x, y)$ of the entire prompt-response can be decomposed into token-wise rewards, i.e., $r(x, y) = \sum_{t=1}^T \gamma^{t-1} R\left([x, y^{<t}], y^t\right)$.
Then, the state-action value function and state value function can be defined as:
\begin{equation}   \label{Equation: advantage function Q}
    \begin{aligned}
        Q_{\pi} \left([x, y^{<t}], y^t\right)
         & =\mathbb{E}_{\pi} \left[\left. \sum_{m=0}^{T-t}\gamma^m R_{t + m}\right|[x, y^{<t}], y^t \right], 
    \end{aligned}
\end{equation}
\begin{equation}   \label{Equation: advantage function V}
    \begin{aligned}
        V_{\pi} \left([x, y^{<t}] \right)
         & =\mathbb{E}_{z\sim\pi(\cdot \mid [x, y^{<t}])}\left[Q_\pi([x, y^{<t}], z)\right].
    \end{aligned}
\end{equation}
The advantage function can be expressed as follows:
\begin{equation} 
A_{\pi} \left([x, y^{<t}], y^t\right) = Q_{\pi} \left([x, y^{<t}], y^t \right) - V_{\pi}\left([x, y^{<t}] \right).
\end{equation}

Since token generation is formulated as an undiscounted finite-horizon process, we set $\gamma=1$. 
Following TDPO~\citep{zeng2024token}, we consider the following KL-regularized token-level objective:
\begin{equation}
    \begin{aligned}
        \max_{\pi_\theta}\; &\mathbb{E}_{\substack{ (x,y^{<t})\sim\mathcal D\\
        a\sim\pi_\theta(\cdot\mid[x,y^{<t}])}} \Big[A_{\pi_{\mathrm{ref}}}([x,y^{<t}],a) \\
        &\quad -\beta D_{\mathrm{KL}}\big(\pi_\theta(\cdot\mid[x,y^{<t}]) \| \pi_{\mathrm{ref}}(\cdot\mid[x,y^{<t}]) \big)\Big].
    \end{aligned}
    \label{eq:dro-dpo}
\end{equation}
Here, $a$ denotes the next generated token and $\pi_{\mathrm{ref}}$ is the fixed reference policy.

Then, based on Lemma 4.5 in \cite{zeng2024token}, we can establish the equivalence between the BT model and the Regret Preference Model as shown in the following lemma.

\begin{lemma} \label{BT of TDPO}
    Given a prompt $x$ and pairwise responses $\left(y_c, y_r\right)$, the Bradley-Terry model expresses the preference probability in terms of the goal-conditioned optimal policy $\pi_\theta^*$ and reference policy $\pi_{\mathrm{ref}}$:
    \begin{equation} 
        P_{\mathrm{BT}}^{\ast}\left(y_c \succ y_r \mid x\right) = \sigma\left(\Delta M^{\ast} - \Delta K^{\ast}\right)
    \end{equation}
    where $\Delta M$ is the difference in implicit rewards defined in Eq.~\eqref{eq:delta_m}, $\Delta K$ represents the difference in sequential KL divergence between two pairs $\left(x, y_c\right)$ and $\left(x, y_r\right)$, expressed as:
    \begin{equation}
        \begin{aligned}
        \Delta K = & \beta D_{\mathrm{SeqKL}} \left(x, y_r; \pi_{\mathrm{ref}} \| \pi_{\theta}\right) \\
        & -\beta D_{\mathrm{SeqKL}}\left(x, y_c; \pi_{\mathrm{ref}} \| \pi_{\theta}\right).
        \end{aligned}
    \end{equation}
The sequential KL divergence is defined in Appendix~A.2.
\end{lemma}

Drawing on Lemma~\ref{BT of TDPO}, we reformulate the BT model into a structure solely relevant to the policy, which enables us to formulate a likelihood maximization objective for a parametrized policy $\pi_{\theta}$.
The loss function is given by:
\begin{equation} \label{Equation: BTDPO loss}
    \min_{\theta} \max_{\mathbf{P}\in\mathcal{P}(\rho;\mathbf{P}^{o})} -\mathbb{E}_{z\sim\mathbf{P}} \left[\log \sigma\left(\Delta M - \Delta K\right) \right].
\end{equation}
where $\mathbf P^\circ$ is the empirical preference-sample
distribution and
\begin{equation}
\mathcal P(\rho;\mathbf P^\circ)
=
\left\{
\mathbf P:
D_{\mathrm{KL}}(\mathbf P\|\mathbf P^\circ)\leq\rho
\right\}.
\end{equation}

\paragraph{VLM-Guided RL.}

In the online stage, we use the fine-tuned VLM as a goal planner to guide the RL agent.
At each time step $t$, the VLM planner receives the query $x_t=(o_t,d_t,l)$, where $o_t$ is the current visual observation, $d_t$ is the textual description of the observation, and $l$ is the task instruction.
Then, the VLM planner generates a prioritized goal set
$ y_t = \{(g_t^k,\rho_t^k)\}_{k=1}^{K},$
where $g_t^k$ is the $k$-th candidate goal, and $\rho_t^k$ is its priority score.
The RL agent selects the action according to $a_t \sim \pi_\phi(\cdot\mid o_t,y_t^{\mathrm{emb}})$ to maximize the expected sum of discounted rewards, where $y_t^{\mathrm{emb}}$ is the text embedding of the prioritized goal set.

The policy is optimized using PPO~\citep{schulman2017proximal} with goal-conditioned state augmentation.
Here, the probability ratio is defined as
\begin{equation}
    \eta_t(\phi)=\frac{\pi_\phi(a_t\mid o_t,y_t^{\mathrm{emb}})}{\pi_{\phi_{\mathrm{old}}}(a_t\mid o_t,y_t^{\mathrm{emb}})},
\end{equation}
where $\pi_{\phi_{\mathrm{old}}}$ is the policy before the update.
The clipped PPO surrogate objective is
\begin{equation}
    \begin{aligned}
        & \mathcal{L}_{\mathrm{PPO}}(\phi)
        = \\
        & \mathbb{E}_t
        \Bigg[
        \min
        \left(
        \eta_t(\phi)\hat{A}_t,
        \mathrm{clip}(\eta_t(\phi),1-\epsilon,1+\epsilon)\hat{A}_t
        \right)
        \Bigg],
    \end{aligned}
    \label{eq:ppo_objective}
\end{equation}
where $\hat{A}_t$ is the estimated advantage function and $\epsilon$ is the clipping coefficient.

\section{Experiments}

We conduct experiments on the open-world environment Craftax-Classic~\citep{matthews2024craftax}.
Our experiments aim to assess whether the fine-tuned VLM can generate effective goals that guide the RL agent toward efficient exploration and improved task performance.

\subsection{Experimental Setup}

\paragraph{Environment.}

Craftax-Classic is a JAX-based reimplementation of Crafter~\citep{hafner2022benchmarking} with open-world survival and crafting mechanics~\citep{matthews2024craftax}.
It features a procedurally generated 2D world in which the agent explores, gathers resources, crafts tools, combats hostile entities, and unlocks achievements.
The environment includes 22 achievements, $63\times63\times3$ visual observations, and 17 discrete actions, providing a challenging benchmark for goal-guided exploration and long-horizon planning.
Further details about the environment can be found in Appendix~C.1.

\paragraph{Baselines.}

We compare against three groups of baselines:
conventional RL without language-model guidance (PPO~\citep{schulman2017proximal});
cloud-based LLM-guided methods (ELLM~\citep{du2023guiding} and AdaRefiner~\citep{zhang2024adarefiner}), which use DeepSeek-V3 as the planner;
and methods with locally deployed planners (Causal-aware LLMs~\citep{chen2025causal} and SGRL~\citep{qi2025goal}).
Causal-aware LLMs use the same base VLM as our method without fine-tuning.
For SGRL, we report results using planning code generated by Qwen3-Coder-30B-A3B-Instruct and DeepSeek-R1, respectively.
We also compare planners based on the base VLM and versions fine-tuned with SFT, DPO~\citep{rafailov2023direct_dpo}, TDPO~\citep{zeng2024token}, and KLDPO~\citep{xu2025robust}.

\paragraph{Training Configuration.}

We use Qwen3-VL-8B-Instruct\footnote{\url{https://huggingface.co/Qwen/Qwen3-VL-8B-Instruct}} as the base model and compute the goal-text embeddings using \texttt{paraphrase-MiniLM-L6-v2}\footnote{\url{https://huggingface.co/sentence-transformers/paraphrase-MiniLM-L6-v2}}.
All VLM fine-tuning experiments are conducted on four NVIDIA H100 GPUs with 80 GB of memory each.
Detailed PPO hyperparameters and VLM training configurations are provided in Appendix~C.1.

\paragraph{Evaluation Metrics.}

We evaluate RL agent performance using three metrics: cumulative reward, overall score and achievement success rate.
The cumulative reward measures achievement completion and survival feedback, with the agent receives $+1$ for each newly unlocked achievement and $+0.1$ or $-0.1$ for recovering or losing health points.
The overall score aggregates achievement difficulties using the log-space average:
$S \doteq \exp \left(\frac{1}{N} \sum_{i=1}^{N} \ln(1+s_i) \right)-1,$
where $s_i\in[0;100]$ is the success rate of the $i$-th achievement and $N=22$.
The success rate of each achievement is computed as the proportion of episodes in which the achievement is unlocked.

\begin{figure*}[t]
\centering
\includegraphics[width=0.85\linewidth]
{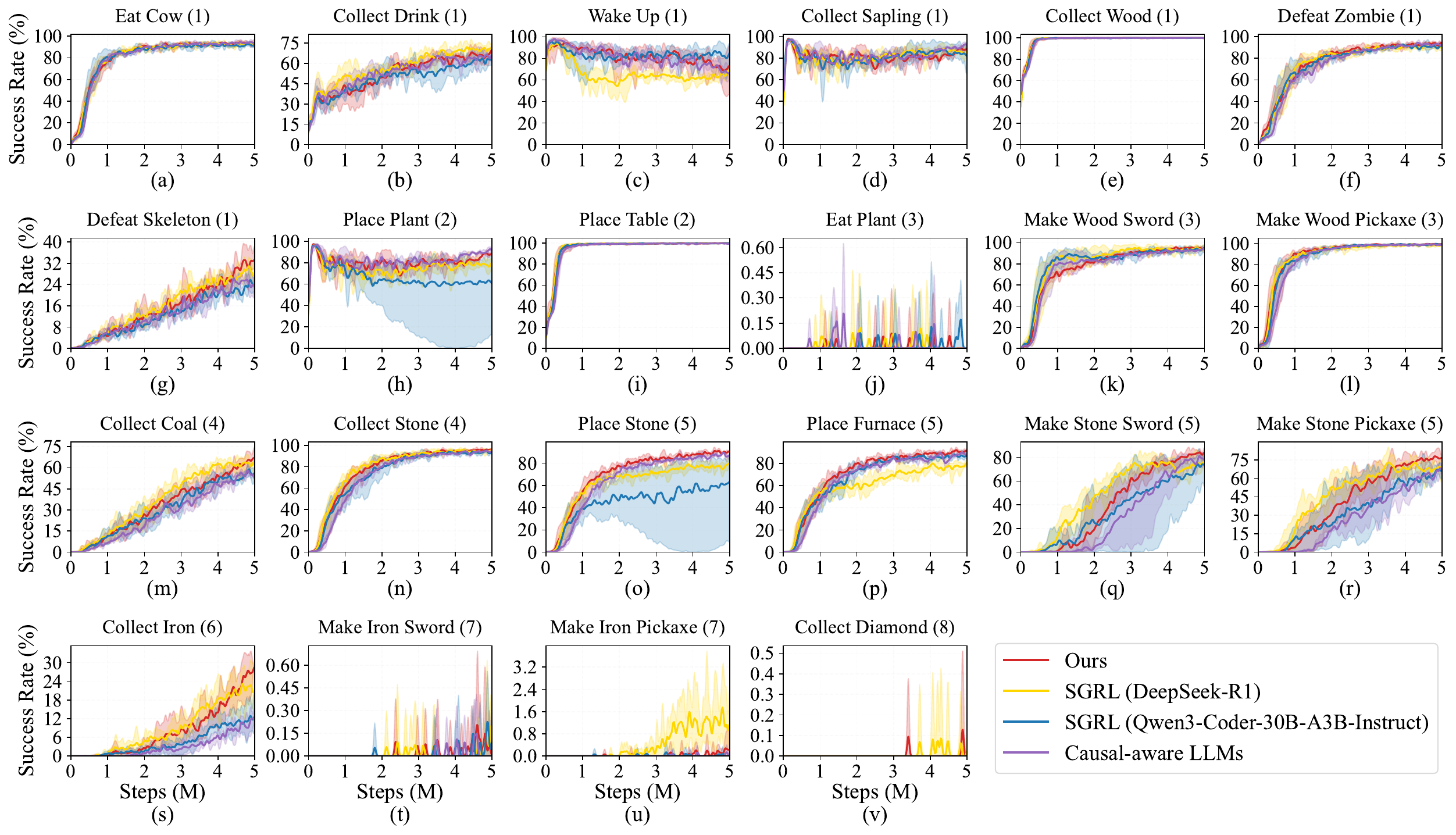}
\caption{Success rate curves for all 22 achievements (ranked by depth and importance) on Craftax-Classic. All results are the mean (solid lines) and standard deviation (shaded areas) over five seeds.}
\label{fig:curves_kldpo_sgrl}
\end{figure*}

\begin{figure*}[t]
\centering
\includegraphics[width=0.85\textwidth]{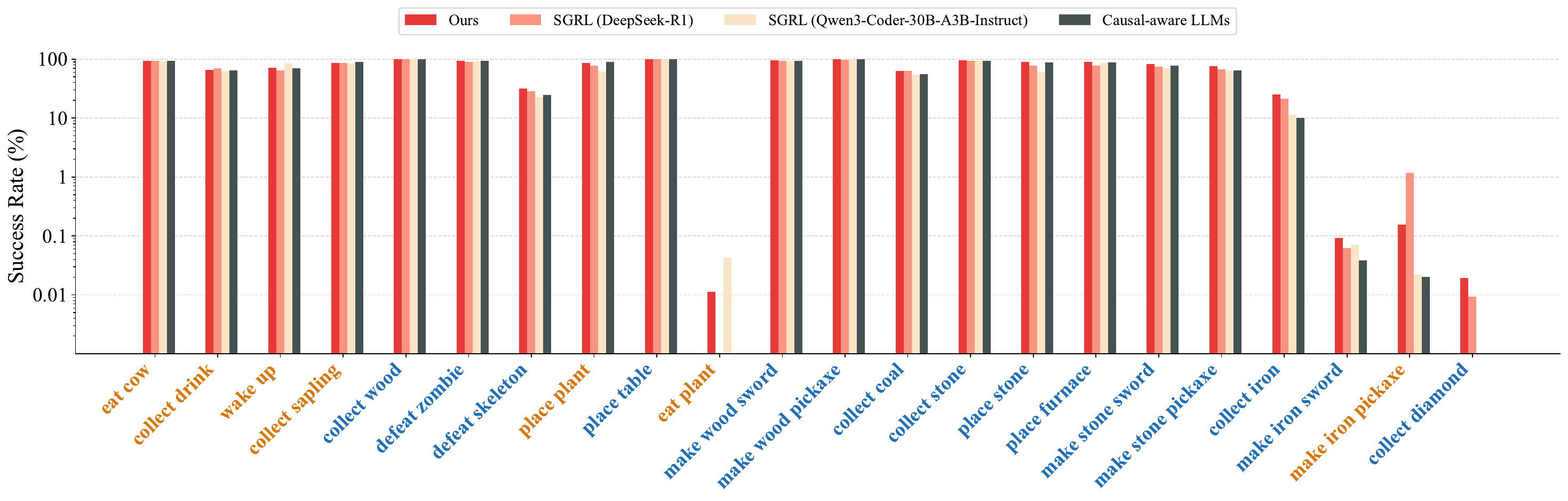} 
\caption{Success rates across all 22 achievements on Craftax-Classic at 5M steps. 
Blue and orange achievement names denote the tasks where our method ranks first and second, respectively.}
\label{fig:main_result}
\end{figure*}

\begin{figure*}[t]
\centering
\includegraphics[width=0.9\textwidth]{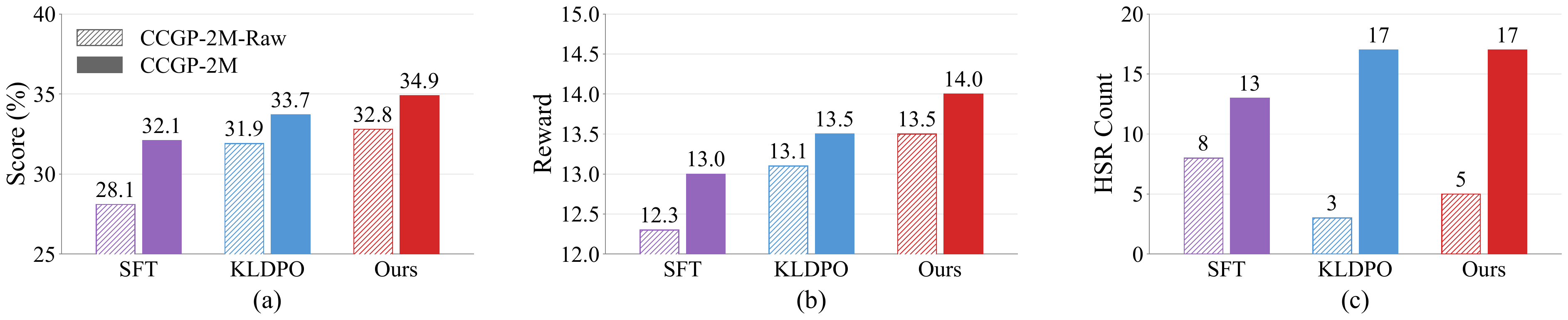} 
\caption{The scores, rewards, and highest success rate (HSR) counts of RL agents guided by different VLMs on the CCGP-2M-Raw and CCGP-2M datasets at 5M steps.}
\label{fig:dataset_comparison2}
\end{figure*}

\begin{table}
\centering
\caption{Performance on Craftax-Classic (mean $\pm$ std) at 5M steps.}
\begin{tabular*}{\linewidth}{@{\extracolsep{\fill}}lcc@{}}
\toprule
\textbf{Method} & \textbf{Score (\%)} & \textbf{Reward} \\
\midrule
PPO~\citep{schulman2017proximal}              & $24.8 \pm 5.7$ & $11.9 \pm 1.1$ \\
ELLM~\citep{du2023guiding}              & $28.4 \pm 2.5$ & $12.2 \pm 1.0$ \\
AdaRefiner~\citep{zhang2024adarefiner}        & $28.5 \pm 2.3$ & $12.3 \pm 0.9$ \\
Causal-aware LLMs~\citep{chen2025causal} & $31.8 \pm 0.5$ & $12.9 \pm 0.4$ \\
\makecell[c]{SGRL (Qwen3-Coder-30B-\\A3B-Instruct) \citep{qi2025goal}}
  & $30.6 \pm 2.7$ & $12.8 \pm 0.6$ \\
SGRL (DeepSeek-R1)~\citep{qi2025goal}              & $33.8 \pm 1.5$ & $13.0 \pm 0.3$ \\
\midrule
Base VLM          & $26.7 \pm 3.3$ & $11.8 \pm 0.5$ \\
SFT               & $32.1 \pm 0.7$ & $13.0 \pm 0.2$ \\
DPO               & $31.9 \pm 1.1$ & $13.1 \pm 0.3$ \\
TDPO              & $32.6 \pm 1.0$ & $13.4 \pm 0.2$ \\
KLDPO             & $33.7 \pm 1.0$ & $13.5 \pm 0.3$ \\
KLTDPO (\textbf{Ours})    & $\mathbf{34.9 \pm 0.7}$ &
                     $\mathbf{14.0 \pm 0.2}$ \\
\bottomrule
\end{tabular*}
\label{tab:main_results}
\end{table}

\subsection{Main Results}

We evaluate the proposed method on Craftax-Classic using score, reward, and achievement success rates.
Table~\ref{tab:main_results} reports the mean and standard deviation of the score and reward on Craftax-Classic at 5M steps.
PPO achieves the lowest score, while the language-guided baselines improve upon it.
Among these baselines, SGRL with DeepSeek-R1-generated planning code achieves the highest score. 
For VLM-based planners, SFT substantially improves performance over the base VLM, and TDPO and KLDPO further improve both score and reward over SFT.
KLTDPO achieves the highest mean score and reward among all evaluated methods, with relatively small standard deviations across runs.
These results suggest that KLTDPO provides effective goal guidance and improves downstream RL performance at 5M steps.

Fig.~\ref{fig:curves_kldpo_sgrl} compares learning curves across all 22 achievements. 
Our method reaches higher success rates than SGRL with Qwen3-Coder-generated planning code and Causal-aware LLMs on several intermediate achievements, including placing stone, making stone tools, and collecting iron (see Fig.~\ref{fig:curves_kldpo_sgrl}(o)--(s)). 
Compared with SGRL using DeepSeek-R1-generated planning code, our method remains competitive across many achievements and reaches higher success rates on several tasks from approximately 4.5M to 5M steps (see Fig.~\ref{fig:curves_kldpo_sgrl}(g), (m)--(s)).
Notably, our method first collects diamonds at $3.38$M environment steps, earlier than SGRL with DeepSeek-R1-generated planning code, which first collects diamonds at $3.71$M steps.

Fig.~\ref{fig:main_result} presents the success rates for all 22 achievements at 5M steps. 
Among the methods compared in the figure, our method achieves the highest success rate on 15 achievements and the second-highest on the remaining achievements.  
These results show competitive performance across achievements of varying depth, with particularly strong results on several tool-crafting tasks. 
However, success rates on the deepest achievements remain low, indicating that reliable completion of these tasks is still challenging.
Further comparisons with other baselines and VLM fine-tuning methods are provided in Appendix~C.2.

Overall, these results suggest that a planner fine-tuned with KLTDPO on CCGP-2M provides effective goal guidance for downstream RL, achieving competitive performance across a range of achievements.

\subsection{Data Quality Analysis} 

Fig.~\ref{fig:dataset_comparison2} presents the scores, rewards, and highest success rate (HSR) counts of RL agents guided by VLMs fine-tuned with SFT, KLDPO, and KLTDPO on the CCGP-2M-Raw\footnote{CCGP-2M-Raw is a preference dataset constructed by randomly selecting two million samples from the complete preference dataset.} and CCGP-2M datasets, respectively.
Across all three training methods, using CCGP-2M leads to higher scores and rewards than using CCGP-2M-Raw.
These results indicate that removing redundant and low-quality records provides more reliable supervision for VLM fine-tuning and improves downstream RL performance.
More details are provided in Appendix~C.3.

\section{Related Work}

\subsection{LM-Guided RL} \label{sec:lm_guided_rl}

In recent years, large language models (LLMs) have been leveraged to guide RL agents in addressing the challenges of sparse rewards and long-horizon challenges in open-world environments~\citep{Zhou2024LLM4RL,schoepp2025evolving}.
One line of research leverages cloud-based LLMs as external planners to guide RL agents by generating auxiliary rewards or goals~\citep{du2023guiding}, refining decisions from agent feedback~\citep{zhang2024adarefiner}, or producing subgoal sequences and planning programs~\citep{shukla2024lgts,gu2026sto,qi2025goal}. 
However, their reliance on large-scale external models limits local deployment and frequent interaction.
Another line of research explores locally deployable planners using task-specific prompts~\citep{ruiz2024words}, structured planning knowledge such as causal relations and subgoal graphs~\citep{yuan2023skill,chen2025causal,fan2025subgoal}, or self-correction mechanisms~\citep{chi2025llmv,dou2026dsadf}. 
Nevertheless, these approaches mainly enhance general planning capabilities without explicitly learning environment-specific goal preferences.
Recent VLM-based RL methods further incorporate visual-semantic understanding through reward modeling~\citep{jiang2024reinforcement}, trajectory feedback~\citep{luu2025enhancing}, or representation learning~\citep{xu2025vlms}.
Unlike these approaches, we learn environment-specific goal preferences from interaction trajectories and align a local VLM planner through preference optimization for adaptive RL guidance.

\subsection{Preference Optimization}

Beyond SFT, preference-based methods align models by learning from comparisons between preferred and rejected responses.
RLHF first learns a reward model from preference data and then optimizes the policy using RL~\citep{christiano2017deep,ouyang2022training}.
DPO simplifies this process by directly optimizing the policy from preference samples without training an explicit reward model~\citep{rafailov2023direct_dpo}.
Subsequent methods improve DPO from different perspectives.
KLDPO formulates preference optimization under distributional uncertainty and assigns larger weights to preference samples with higher losses, improving robustness to preference shifts~\citep{xu2025robust}.
TDPO introduces token-level sequential KL regularization to control policy deviations at each generation step~\citep{zeng2024token}.
However, KLDPO still evaluates each response as a whole, while TDPO does not account for differences in preference-pair difficulty.
We therefore propose KLTDPO, which combines token-level sequential KL regularization with distributionally robust loss reweighting.
This design is tailored to CCGP-2M, where chosen and rejected plans may share most of their valid content and differ only in a goal, ordering decision, or priority score.

\section{Conclusion}

In this paper, we present a VLM-guided RL framework that learns environment-specific goal guidance from interaction trajectories. 
We construct CCGP-2M, a multimodal dataset containing 2 million goal-preference samples collected in Craftax-Classic. The proposed framework combines offline VLM fine-tuning with online goal-guided RL. 
In the offline stage, KLTDPO aligns the VLM with goal preferences by combining token-level KL regularization with distributionally robust reweighting. 
In the online stage, the fine-tuned VLM generates high-level goals to guide the RL agent's exploration. 
Experiments on Craftax-Classic show that our method collects diamonds earlier, attains the highest score and reward, ranks first on 15 achievements and second on the remaining achievements.
Further analysis shows that quality-aware data selection improves downstream performance.
Together, these results suggest a viable path toward learning environment-specific goal guidance from interaction data to support exploration in open-world RL.
Nevertheless, the deepest achievements remain challenging, and our evaluation is limited to Craftax-Classic. 
Future work will investigate more cost-efficient data collection and training methods, and evaluate the framework in a broader range of environments.

{
    \small
    \bibliographystyle{ieeenat_fullname}
    \bibliography{main}
}


\end{document}